\documentclass[11pt]{article}

\usepackage{section/acl}

\usepackage{times}
\usepackage{latexsym}

\usepackage[T1]{fontenc}

\usepackage[utf8]{inputenc}

\usepackage{microtype}

\usepackage{inconsolata}

\usepackage{graphicx}

\usepackage{enumitem}

\usepackage{booktabs}   
\usepackage{multirow}   
\usepackage{caption}    
\usepackage{float}      

\usepackage{color}
\usepackage{tcolorbox}          
\tcbuselibrary{skins,breakable} 
\usepackage[table]{xcolor}

\definecolor{tred}{RGB}{251, 130, 132}
\newcommand{\highlightbox}[3]{
\begin{tcolorbox}[
    enhanced jigsaw,
    breakable,
    pad at break*=1mm,
    colback=#1!5!white,
    colframe=#1,
    title=#2
]
#3
\end{tcolorbox}
}

\usepackage{tikz}
\usetikzlibrary{shapes.geometric, arrows.meta, positioning, calc, shadows}

\title{FTibSuite: A Comprehensive Resource Suite for Tibetan Vision–Language Modeling}

\author{%
  \small
  \begin{tabular}[t]{@{}c@{}}
    Guixian Xu\textsuperscript{1,2$\ast$} \quad
    Yide Liang\textsuperscript{2$\ast$} \quad
    Zeli Su\textsuperscript{2$\ast\dagger$} \quad
    Xuexian Song\textsuperscript{4$\ast$} \\
    Ziyin Zhang\textsuperscript{3$\ast\dagger$} \quad
    Yushuang Dong\textsuperscript{2} \quad
    Ting Zhang\textsuperscript{2} \quad
    Xu Han\textsuperscript{2}
  \end{tabular}
  \\
  \vspace{2pt}
  \footnotesize
  \begin{tabular}[t]{@{}c@{}}
    \textsuperscript{1}Hainan International College, Minzu University of China \\
    \textsuperscript{2}School of Information Engineering, Minzu University of China \\
    \textsuperscript{3}Shanghai Jiao Tong University \\
    \textsuperscript{4}Institute of Automation, Chinese Academy of Sciences
  \end{tabular}
  \\
  \vspace{2pt}
  \footnotesize\texttt{\{guixian\_xu,24302278,zeli\_su,yushuangdong,hanxu\}@muc.edu.cn} \\
  \footnotesize\texttt{tozhangting@126.com} \quad
  \footnotesize\texttt{ziyinzhang@sjtu.edu.cn} \quad
  \footnotesize\texttt{songxuexian5@gmail.com}
}

\begin{document}
\maketitle
\begin{abstract} 
Vision–language models (VLMs) have progressed rapidly, but Tibetan remains largely underserved due to the lack of infrastructure for reproducible training and evaluation. To help address this gap, we introduce FTibSuite, a resource-centric foundation for Tibetan VLM research that provides an end-to-end training-and-evaluation workflow and includes human-verified multimodal annotations, partially filling a long-standing shortage of Tibetan multimodal resources. FTibSuite comprises FTibData, FTibBench, and a reproducible baseline model, FTibVLM, built on Qwen3-VL-8B-Instruct. FTibVLM adopts a three-stage adaptation pipeline consisting of Tibetan continual pretraining, image–text alignment, and multimodal instruction tuning.
For systematic evaluation, FTibBench adapts five established multimodal benchmarks to Tibetan and offers a reproducible evaluation protocol to support consistent comparisons across models. Specifically, FTibBench includes Tibetan versions of MMBench, MME, POPE, BinaryVQA, and COREVQA. Experiments on FTibBench demonstrate that FTibVLM consistently improves Tibetan multimodal performance. For instance, FTibVLM attains 76.01 accuracy on BinaryVQA, indicating that Tibetan performance can be competitive with high-resource settings on this diagnostic task. We also observe substantial gains on other benchmarks, including an improvement on MMBench from 42.97 to 67.78 and an increase in POPE-random accuracy from 47.53 to 80.56, underscoring the practical value of the proposed workflow and resources.
\end{abstract}

\begin{center} \small \includegraphics[width=1em,height=1em]{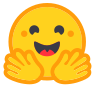}\hspace{0.35em} \begin{tabular}{@{}l@{}} \url{https://huggingface.co/onedday/FTib-VLM}\\ \url{https://huggingface.co/datasets/onedday/FTib-data} \end{tabular} \end{center}

\section{Introduction}
In recent years, large language models (LLMs) and their multimodal extensions, such as vision–language models (VLMs) and multimodal large language models (MLLMs), have achieved remarkable progress in dialogue, reasoning, and visual understanding and generation \cite{achiam2023gpt}. 
These advances have been largely driven by the continuous accumulation of high-quality pretraining data and the maturation of large-scale training paradigms \cite{kaplan2020scaling,hoffmann2022training}. 
However, such capability gains are uneven across languages: high-resource languages like English and Chinese continue to benefit from a positive feedback loop of data availability and ecosystem development, whereas many low-resource languages face the dual constraints of scarce high-quality training data and a lack of standardized, reproducible evaluation protocols \cite{costa2022no,conneau2020unsupervised}. 
This, in turn, hampers effective adaptation and undermines the comparability and reproducibility of different methods.  

\begin{figure}
    \centering
    \includegraphics[width=1\linewidth]{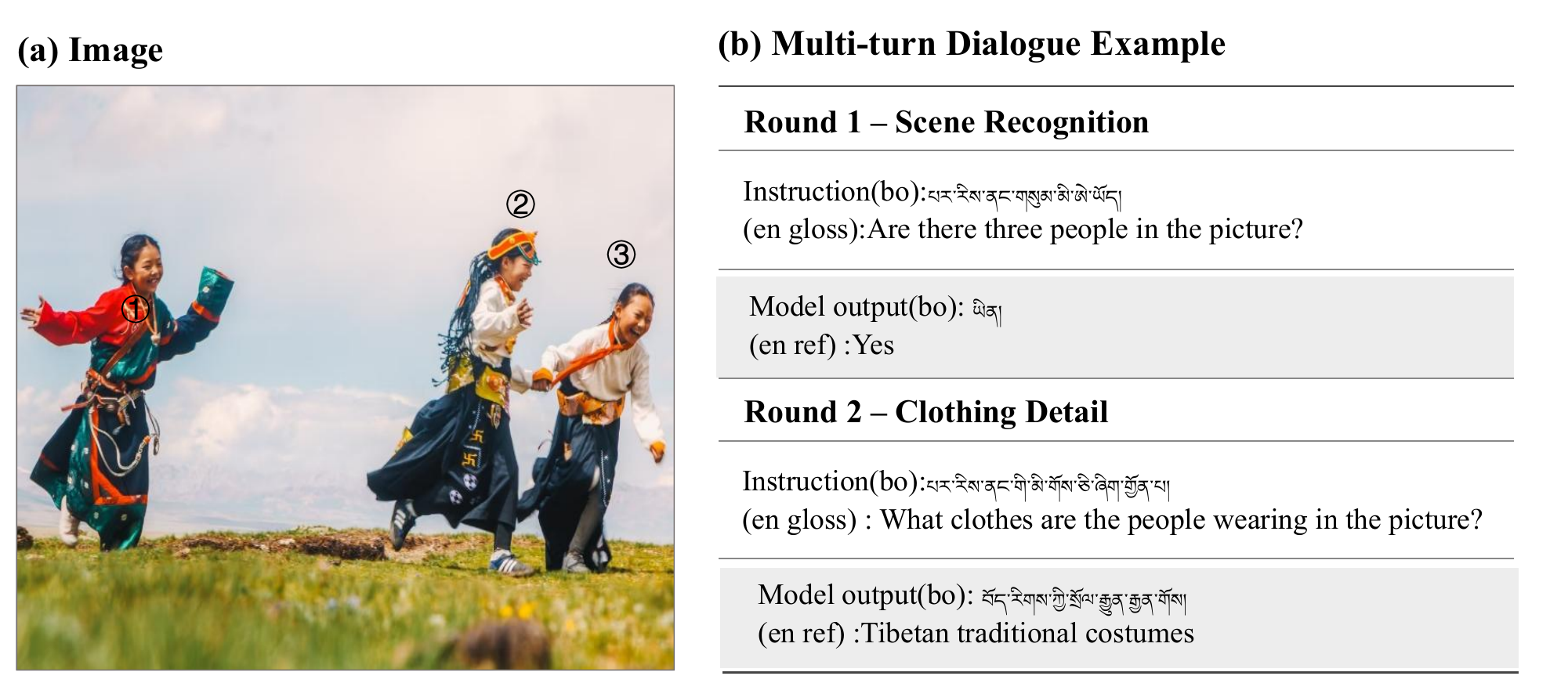}
    \caption{Panel (a) shows the input image featuring traditional Tibetan elements. Panel (b) displays a two-round interaction where the model responds to Tibetan instructions on scene recognition by determining the number of people, and on fine-grained visual understanding by identifying the clothing style. English translations are provided for reference.}
    \vspace{-0.5cm}
    \label{fig:1_plot}
\end{figure}

Tibetan is a typical low-resource language. 
Although progress has been made on certain text-based tasks (e.g., neural machine translation and text classification) \cite{an2023prompt,chen2025adapting}, the scale and quality of publicly accessible data remain markedly insufficient \cite{huang2025tibetan}. 
Meanwhile, differences in the Tibetan writing system and its punctuation/whitespace conventions, as well as frequent code-mixing with Chinese and English, further complicate data processing, making automatic data collection and cleaning more prone to failure \cite{li2024tibetan}. 

In the multimodal era, this challenge is further amplified. 
Building and evaluating VLMs requires not only text corpora, but also large-scale, high-quality image–text alignment data, multimodal instruction datasets, and standardized benchmarks with reproducible experimental setups. 
However, in the Tibetan setting, publicly available resources for image–text alignment, multimodal instruction data, and systematic evaluation protocols remain scarce \cite{alam2025behind}. 
As a result, progress in Tibetan VLM research and reproducibility has been slow, and it is difficult to conduct reliable and fair capability assessments under unified conditions.  Figure~\ref{fig:1_plot} provides a motivating example of Tibetan multi-turn vision-language interaction, where the model is required to follow Tibetan instructions and perform fine-grained visual grounding. This also highlights the need for reliable training signals and standardized evaluation infrastructure tailored to Tibetan.

To fill this longstanding infrastructure gap, we propose a resource suite for Tibetan multimodal research, collectively named \textbf{FTibSuite}. 
FTibSuite consists of three components: 
(i) \textbf{FTibVLM}, a reproducible Tibetan vision--language model baseline built upon the strong open-source backbone \textbf{Qwen3-VL-8B-Instruct}, trained via a general staged adaptation pipeline consisting of continued pretraining, multimodal alignment, and multimodal instruction fine-tuning; 
(ii) \textbf{FTibData}, a Tibetan data collection that supports both training and instruction tuning; and 
 (iii) \textbf{FTibBench}, a high-quality evaluation suite constructed by translating and adapting multiple mainstream multimodal benchmarks to the Tibetan setting, enabling systematic evaluation of Tibetan VLMs. 

Because translating and adapting benchmarks can easily introduce non-trivial systematic noise, the reliability of the evaluation suite is particularly important. To improve the quality of \textbf{FTibBench}, we adopt a hierarchical quality-control pipeline that uses \textbf{DeepSeek-V3} for automatic verification and scoring to identify translation inconsistencies and other high-risk errors, routes low-quality cases to Tibetan-language experts for mandatory correction, and further audits 10\% of the automatically accepted samples through human review. Together, these procedures help reduce systematic bias and provide a more credible foundation for Tibetan multimodal evaluation.

Experimental results show that this reproducible, data-and-benchmark-centric pipeline substantially improves Tibetan multimodal capabilities on top of the backbone baseline, and provides the first comprehensive and reproducible experimental evidence for systematic evaluation of Tibetan VLMs. 

Our contributions are summarized as follows: 
\begin{itemize}[leftmargin=0.5cm,itemsep=0cm,topsep=1pt,parsep=1pt]
\item We release \textbf{FTibVLM}, the first reproducible Tibetan VLM baseline built upon a strong open-source backbone model. 

\item We construct and open-source \textbf{FTibData}, a training data collection covering the key data types required throughout the full adaptation pipeline, including Tibetan text corpora for continual pretraining, Tibetan image--text data for multimodal alignment, and Tibetan instruction data for multimodal instruction fine-tuning. 

\item We build \textbf{FTibBench}, a systematic benchmark suite for Tibetan VLMs, by translating and adapting five widely used multimodal benchmarks, including BinaryVQA and MMBench, to the Tibetan setting, enabling comprehensive evaluation of Tibetan VLMs across diverse capability dimensions. 
\end{itemize}
\section{Related Work}
\subsection{Vision-Language Models}  
The development of vision–language models has been largely driven by two complementary lines of research: large-scale image–text pretraining and unified generative modeling. 
Early work typically learns cross-modal representations from web-scale image–text pairs via contrastive objectives, exemplified by CLIP~\cite{radford2021learning}, ALIGN~\cite{jia2021scaling}, and BLIP~\cite{li2022blip,li2023blip}.

As instruction following has emerged as a de facto interface for LLMs, multimodal research has increasingly shifted toward LLM-centric, generative VLMs. 
For example, Flamingo \cite{alayrac2022flamingo} introduces cross-modal connector modules between vision and language backbones, while PaLI \cite{chen2022pali} emphasizes joint scaling of vision and language. 
In the open-source ecosystem, LLaVA \cite{liu2023visual} and InstructBLIP \cite{dai2023instructblip} demonstrate that converting heterogeneous multimodal tasks into an “instruction–response” format is a key step toward building general-purpose visual assistants, while Qwen-VL \cite{bai2023qwen} systematically highlights the importance of a strong backbone, multi-stage training, and curated multilingual multimodal corpora for general capabilities. 

\subsection{Data and Evaluation for VLMs}
The advancement of VLMs has been largely enabled by the joint maturation of high-quality instruction data and diagnostic evaluation suites. 
For extremely low-resource languages such as Tibetan, the availability of a reusable data pipeline spanning continual pretraining to instruction tuning is often a key determinant of whether an open and sustainable research ecosystem can be established. 

On the data side, instruction construction is increasingly moving beyond purely synthetic QA toward broader task coverage and higher annotation quality. 
Vision-Flan \cite{xu2024vision}, for example, reformulates a wide range of academic datasets into a unified visual instruction format and demonstrates the effectiveness of a two-stage instruction-tuning recipe, first leveraging high-quality human-labeled tasks and then scaling with synthetic alignment data. 
LoResMT \cite{xiao2025text} further explores systematic pipelines that transform parallel text corpora into multimodal training data in low-resource settings.

On the evaluation side, benchmarks are shifting from coarse-grained leaderboards toward diagnostic and reliability-oriented assessments. 
POPE \cite{li2023evaluating} evaluates object hallucination in VLMs. 
MME \cite{fu2025mme}, in contrast, offers a more comprehensive capability profile by covering both perception- and cognition-level sub-tasks. 
Beyond these, BinaryVQA \cite{borji2023binaryvqa} probes out-of-distribution generalization and bias, while COREVQA \cite{chintapatla2025corevqa} targets fine-grained observation and reasoning in crowded scenes, further revealing the brittleness of current VLMs under challenging visual conditions. 
More recently, UPD \cite{miyai2025unsolvable} highlights that high multiple-choice VQA scores alone do not necessarily imply genuine understanding.
However, despite the abundance of existing evaluations, they are predominantly English-centric, and there is no widely adopted, publicly released Tibetan counterpart of mainstream multimodal benchmarks. 

\begin{figure*}[!t]
\begin{center}
\includegraphics[width=0.9\textwidth,keepaspectratio]{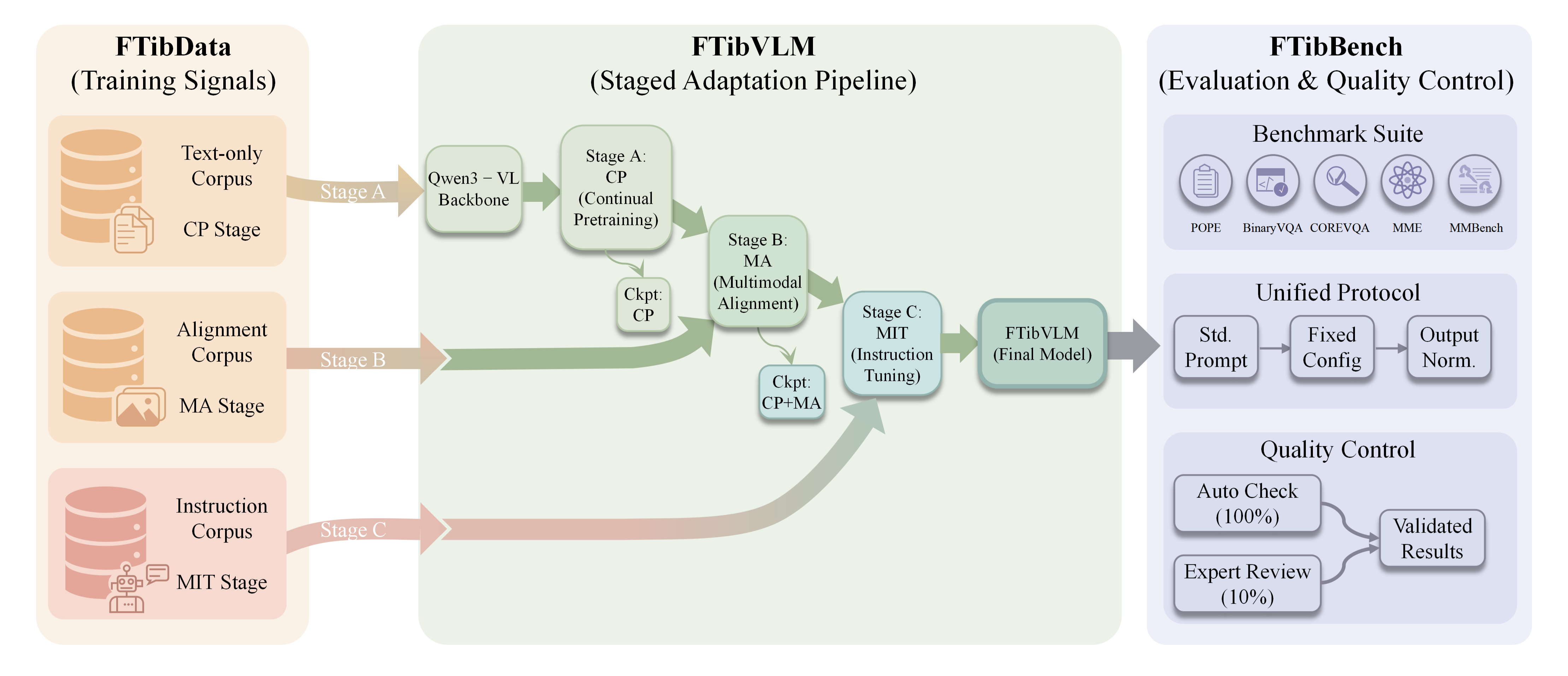}
\end{center}
\vspace{-0.4cm}
\caption{\textbf{FTibSuite overview.} It consists of three coupled components: FTibData, which provides reusable multilingual and multimodal training signals; FTibVLM, a staged adaptation pipeline that incrementally adapts a vision--language backbone to Tibetan via continual pretraining (CP), multimodal alignment (MA), and multimodal instruction tuning (MIT); and FTibBench, a unified evaluation framework with standardized protocols and hierarchical quality control.
}
\vspace{-0.3cm}
\label{fig:structure}
\end{figure*}

\subsection{Low-Resource Language Adaptation and Resources}
Low-resource language capability is typically achieved through strong backbone plus target-distribution adaptation strategy, such as continual pretraining \cite{gururangan2020don}.
The same principle applies in the multimodal setting: rather than training from scratch, it is often more effective and cost-efficient to start from a strong multimodal backbone and perform distribution alignment and stage-wise adaptation. 

For adaptation efficiency and stability, parameter-efficient fine-tuning provides a solution for low-resource and multi-stage training. 
Adapters enable multi-task expansion by inserting lightweight modules while keeping the backbone frozen \cite{houlsby2019parameter}, and LoRA~\citep{hu2022lora} substantially reduce memory and parameter overhead through low-rank updates and quantized training, making iterative development feasible under limited compute budgets. 
BranchLoRA \cite{zhang2025enhancing} further mitigates catastrophic forgetting in continual learning via structured routing and freezing mechanisms. 
Taken together, these studies suggest that effective low-resource adaptation should not only acquire new capabilities, but also preserve existing ones and support controllable transfer. 

From the perspective of resource development, there has been notable progress on Chinese minority languages in terms of textual corpora and pretrained models. 
MC$^2$ \cite{zhang2023mc} systematically constructs multilingual corpora for minority languages in China, while CINO \cite{yang2022cino} and XLM-SWCM~\citep{XLM-SWCM} train dedicated multilingual language models for Chinese minority languages. 
However, these efforts primarily focus on text-only resources. Publicly available multimodal alignment data, instruction-tuning data, and standardized evaluation pipelines for Tibetan remain absent. 

\section{Constructing a Comprehensive VLM Suite for Tibetan}

This section builds and releases the first comprehensive Tibetan research resource and evaluation infrastructure suite for vision--language models (VLMs), \textbf{FTibSuite}, for the Tibetan community. It aims to address three long-standing foundational gaps in Tibetan multimodal research: (i) the lack of reusable training corpora, (ii) the lack of Tibetan evaluation benchmarks that are aligned with mainstream English benchmarks and whose quality can be verified, and (iii) the lack of baseline models that are reproducible and comparable under a unified evaluation protocol.

To this end, we organize the resources produced in this work following a “data–evaluation–baseline” structure, as summarized in Figure~\ref{fig:structure}. \textbf{FTibData} provides reusable training signals for staged adaptation; \textbf{FTibVLM} instantiates these signals into reproducible reference checkpoints; and \textbf{FTibBench} standardizes evaluation with a unified protocol accompanied by a hierarchical quality-control workflow, whose feedback is used to refine translation and parsing rules for subsequent iterations.

\subsection{Training corpus} 
To introduce stable Tibetan generation capabilities without modifying the model architecture, we first conduct Tibetan-oriented continual pretraining. 
The goal of this stage is to explicitly shift the backbone’s language distribution toward the Tibetan text space, enabling more reliable Tibetan language modeling and providing a solid linguistic foundation for subsequent multimodal alignment and instruction fine-tuning. 

We use three categories of text data: a Tibetan subset from MC$^2$\cite{zhang2024mc2}, publicly available Tibetan instruction data (e.g., tibetan-mix-instruction-tuning-60K), and the Chinese LCSTS corpus. 
After unified cleaning, the combined dataset contains approximately 2.2 million samples, with about 70\% Tibetan and 30\% Chinese. 
Under the low-resource setting, we retain a certain proportion of Chinese data for two main reasons: first, preserving the backbone’s original Chinese capability is practically valuable; second, we interleave source-language data during cross-lingual continual pretraining as data replay, motivated by discussions on mitigating catastrophic forgetting in continual pretraining.\cite{zheng2024breaking}

We construct the cross-modal image--text alignment corpus based on the Chinese captioning data of AI Challenger \cite{AI-Challenger}, using a fixed pool of 100k images.
In total, we build 150k one-image--one-caption pairs: 100k Tibetan pairs translated from primary Chinese captions as the main grounding signal, 30k original Chinese pairs retained to stabilize training, and 20k additional Chinese pairs with alternative captions to enhance expression diversity. A supplementary quantitative validation of translation quality on the FTibData caption subset used for multimodal alignment is provided in Appendix~\ref{app:FTibData}.

We build the multimodal instruction fine-tuning corpus based on \textbf{Vision-Flan} \cite{xu2024vision}, with fixed task-type ratios (caption 25\%, VQA 40\%, classification 20\%, counting 5\%, others 10\%).
We translate 30k sampled instances into Tibetan as the primary instruction set, and translate another 10k instances (with the same ratios) into Chinese to maintain Chinese capability.
We further create a 10k Tibetan--Chinese parallel subset by translating the same image-conditioned instances into both languages, and normalize all data into a unified multimodal instruction--response format.



\subsection{Tibetan visual-linguistic baseline} 

We build \textbf{FTibVLM} on top of the multimodal backbone \textbf{Qwen3-VL-8B-Instruct} ~\cite{yang2025qwen3technicalreport} and adopt a unified conversational interface of ``image + textual instruction'' for both training and inference. The adaptation process follows a three-stage pipeline driven by the three data modules in \textbf{FTibData}.

\paragraph{Stage A — Continual Pretraining (CP).}
This stage adapts the backbone’s language distribution toward Tibetan through text-only continual pretraining. The goal is to establish a stable linguistic foundation for Tibetan generation and understanding, while retaining the backbone’s original Chinese-language performance boundary.

\paragraph{Stage B — Multimodal Alignment (MA).}
Given the linguistic prior obtained in CP, this stage performs caption-based image--text alignment to strengthen the correspondence between visual semantics and Tibetan expressions. This improves cross-modal grounding stability when the model is prompted in Tibetan.

\paragraph{Stage C — Multimodal Instruction Tuning (MIT).}
The final stage fine-tunes the model on multimodal instruction data to enhance instruction following, multi-task execution, and interactive usability. Beyond task accuracy, MIT stabilizes response format and decision behaviors under Tibetan prompts.

To support controlled analysis, we save checkpoints after each stage and evaluate three variants under the same backbone starting point: \textbf{Base} (no Tibetan adaptation), \textbf{CP+MA}, and \textbf{CP+MA+MIT} (the final FTibVLM). These variants correspond one-to-one to the three stages above and together constitute the full staged adaptation pipeline.

\subsection{Benchmarks and Metrics}

FTibBench is designed to address the lack of widely adopted and publicly released multimodal benchmarks for Tibetan. Direct translation from English benchmarks often introduces systematic noise (e.g., negation mismatch, numerical drift, entity misalignment, and option–answer mapping errors), which undermines the reliability of evaluation results. Rather than merely localizing existing datasets, FTibBench aims to provide a reproducible and auditable evaluation protocol with controlled differences across models. The full judging prompt and evaluation policy are provided in Appendix~\ref{app:judge-prompt}.

\paragraph{Benchmark Suite.}
FTibBench covers five major multimodal benchmarks in Tibetan: \textbf{POPE} (random / popular / adversarial subsets), \textbf{BinaryVQA}, \textbf{COREVQA}, \textbf{MME}, and \textbf{MMBench-dev}. FTibBench refers to a unified Tibetan-adapted benchmark suite rather than a single original dataset. It integrates Tibetan adaptations of several established multimodal benchmarks under a common evaluation protocol and quality-control workflow. During translation and adaptation, we preserve the original task definitions and answer spaces as much as possible, and expose all benchmarks through a unified execution interface to facilitate reuse, extension, and controlled comparison. Details about FTibBench, including the mapping between each FTibBench subset and its source benchmark, can be found in Appendix~\ref{app:FTibBench} and Table~\ref{tab:ftibbench_mapping}

\paragraph{Evaluation Protocol.}
To ensure comparability across models, FTibBench standardizes evaluation along three components: prompt formatting, decoding configuration, and output normalization. All models are evaluated under an identical Tibetan prompt template and a fixed inference configuration. For classification-style tasks, we restrict the answer space to reduce ambiguity: multiple-choice benchmarks require outputting only the option symbol, while binary tasks are normalized to a \texttt{0/1} decision space. Outputs are subsequently processed through a unified normalization and parsing procedure, and we additionally report the proportion of unmappable outputs as a stability indicator; in our experiments, this invalid rate is \texttt{0}.

\paragraph{Annotation and Quality Control.}
To improve the credibility of translated benchmarks, we adopt a hierarchical quality-control workflow. For each benchmark, all instances are first screened via automated consistency checking. Low-scoring or high-risk samples are routed to mandatory human correction, while an additional 10\% of automatically accepted samples are audited by Tibetan-language experts as a final quality check. Automated verification follows a unified rubric scoring accuracy (\texttt{0--2}), completeness (\texttt{0--2}), and Tibetan linguistic naturalness (\texttt{0--1}), yielding a traceable quality score in \texttt{0--5}. We conducted small-scale comparative tests across multiple large models together with Tibetan experts, and selected \textbf{DeepSeek-V3}~\cite{deepseekai2025deepseekv32pushingfrontieropen} as the primary automatic verifier due to its stability in Tibetan semantic judgment. Manual inspection focuses on high-risk error categories (entity alignment, negation, numerics, option–answer mapping), and all confirmed issues are fed back to refine translations and parsing rules. The LLM-judge prompt can be found in Appendix \ref{app:judge-prompt}.

Taken together, FTibBench provides not only Tibetan counterparts of mainstream multimodal benchmarks, but also a controlled and auditable evaluation protocol, enabling fair, reproducible, and stability-aware comparison of Tibetan VLMs.

\section{Experiments}
\subsection{Experimental Setup} 
\paragraph{Model Setup.} We use \textbf{Qwen3-VL-8B-Instruct} ~\cite{yang2025qwen3technicalreport} as the backbone model, adopt the same three-stage adaptation pipeline as mainstream open-source VLMs, and keep the model architecture unchanged. 
The three training stages include continual pretraining, cross-modal alignment training, and multimodal instruction fine-tuning. 
Our key motivation for choosing a strong backbone is that its general visual understanding and instruction-following capabilities provide a higher starting point for transferring to low-resource languages, allowing us to focus our primary efforts on completing the Tibetan data and evaluation pipeline rather than training an entire multimodal system from scratch. 

\begin{table*}[!t]
    \footnotesize
    \fontsize{10pt}{10pt}\selectfont
    \centering
    \setlength{\tabcolsep}{3pt}
    \renewcommand{\arraystretch}{1.2}
    \begin{tabular}{l cc cc cc cc cc}
        \toprule
        \multirow{2}{*}{\textbf{Model}} 
        & \multicolumn{2}{c}{\textbf{POPE(random)}} 
        & \multicolumn{2}{c}{\textbf{POPE(popular)}}
        & \multicolumn{2}{c}{\textbf{POPE(adversarial)}} 
        & \multicolumn{2}{c}{\textbf{BinaryVQA}} 
        & \multicolumn{2}{c}{\textbf{COREVQA}} \\[0.5ex]
        \cmidrule(lr){2-3}\cmidrule(lr){4-5}\cmidrule(lr){6-7}\cmidrule(lr){8-9}\cmidrule(lr){10-11}
        & \textbf{Acc} & \textbf{F1}
        & \textbf{Acc} & \textbf{F1}
        & \textbf{Acc} & \textbf{F1}
        & \textbf{Acc} & \textbf{F1}
        & \textbf{Acc} & \textbf{F1} \\[0.5ex]
        \midrule
        Base & 47.53 & 43.62 & 46.38 & 60.65 & 46.22 & 60.49 & 54.46 & 53.08 & 31.49 & 42.16 \\[0.5ex]
        \textbf{FTibVLM(ours)} & \textbf{80.56} & \textbf{80.51} & \textbf{81.70} & \textbf{73.40} & \textbf{78.63} & \textbf{78.49} & \textbf{76.01} & \textbf{73.25} & \textbf{50.85} & 35.52 \\[0.5ex]
        \bottomrule
    \end{tabular}
    \caption{
     Main results on Tibetan hallucination robustness and binary VQA benchmarks.
    }
    \label{table1}
\end{table*}

\begin{table*}[!t]
    \footnotesize
    \fontsize{10pt}{10pt}\selectfont  
    \centering
    \renewcommand{\arraystretch}{1.2}
    \begin{tabular}{l cc cc cc cc cc}
        \toprule
        \multirow{2}{*}{\textbf{Model}} 
        & \multicolumn{2}{c}{\textbf{existence}} 
        & \multicolumn{2}{c}{\textbf{color}}
        & \multicolumn{2}{c}{\textbf{posters}} 
        & \multicolumn{2}{c}{\textbf{scene}} 
        & \multicolumn{2}{c}{\textbf{count}} \\[0.5ex]
        \cmidrule(lr){2-3}\cmidrule(lr){4-5}\cmidrule(lr){6-7}\cmidrule(lr){8-9}\cmidrule(lr){10-11}
        & \textbf{Acc} & \textbf{Acc+}
        & \textbf{Acc} & \textbf{Acc+}
        & \textbf{Acc} & \textbf{Acc+}
        & \textbf{Acc} & \textbf{Acc+}
        & \textbf{Acc} & \textbf{Acc+} \\[0.5ex]
        \midrule
        Base    & 50.00 & 33.33 & 55.00 & 10.00 & 63.95 & 34.69 & 60.25 & 31.50 & 50.00 & 0.00 \\[0.5ex]
       \textbf{FTibVLM(ours)}     & \textbf{88.33} & \textbf{80.00} & \textbf{78.33} & \textbf{63.33} & \textbf{77.55} & \textbf{59.18} & \textbf{75.75} & \textbf{53.00} & \textbf{66.70} & \textbf{33.33} \\[0.5ex]
        \bottomrule
    \end{tabular}
    \caption{\label{fewshot}
     Main results on MME for Tibetan multimodal capability profiling. 
    }
    \label{table2}
\end{table*}

\begin{table*}[!t]
\fontsize{10pt}{10pt}\selectfont  
  \centering
  \renewcommand{\arraystretch}{1.2}
  \begin{tabular}{c |c| c c c c c c}
    \toprule
    \textbf{Model} & \textbf{Overall} & \textbf{LR} & \textbf{AR} & \textbf{RR} & \textbf{FP-S} & \textbf{FP-C} & \textbf{CP} \\[0.5ex]
    \midrule
    Base & 42.97 & 52.57 & 40.14 & 38.01 & 43.54 & 42.00 & 43.41\\[0.5ex]
    \textbf{FTibVLM(ours)} & \textbf{67.78} & \textbf{61.65} & \textbf{63.23} & \textbf{66.07} & \textbf{65.16} & \textbf{68.29} & \textbf{76.01}\\[0.5ex]
    \bottomrule
  \end{tabular}
  \caption{\label{ablation-single}
    Main results on MMBench for Tibetan multimodal understanding and reasoning. 
  }
  \label{table3}
\end{table*}
\paragraph{Implementation details.} All three training stages use parameter-efficient fine-tuning with LoRA, are run in bf16 precision, and are trained with DDP on 8$\times$ RTX 4090 GPUs. 
We use the AdamW optimizer, adopt a cosine learning-rate schedule, and set the gradient clipping threshold to 1.0. 
To match the training budget with the objectives of each stage, we apply stage-specific configurations of frozen and trainable modules. 
In the continual pretraining stage, which focuses on language-distribution adaptation, we freeze the visual encoder and the multimodal projection layer, and inject LoRA only into the language components to enable low-cost transfer. 
In the cross-modal alignment and instruction fine-tuning stages, which target visual alignment and improved instruction-following ability, we freeze the visual encoder while keeping the projection layer trainable, so that we can stably preserve the backbone's visual representations while more effectively adapting the cross-modal mapping and instruction behaviors.
\textbf{Training and hyperparameter details are provided in Appendix~\ref{app:ftibvlm}}.

\paragraph{Benchmarks.} This paper proposes \textbf{FTibBench} to evaluate models' Tibetan multimodal capabilities, covering \textbf{POPE} (random, popular, adversarial), \textbf{BinaryVQA}, \textbf{COREVQA}, \textbf{MME}, and \textbf{MMBench} (all in Tibetan; each benchmark follows the official default data split, or uses the official dev set). Chinese capability retention is reported on \textbf{MMBench-CN}. 
POPE, BinaryVQA, and COREVQA report Accuracy and F1. 
MME strictly follows the official evaluation procedure and reports Acc as well as the stricter Acc+. 
MMBench and MMBench-CN use standard multiple-choice evaluation and report the overall score. 

\paragraph{Evaluation protocols.} All experiments strictly adhered to a unified standard, including Tibetan prompt templates, deterministic decoding settings, a constrained output space, and unified parsing rules. 
Specifically, multiple-choice questions required the model to output only the letter of the option; the binary classification task normalized the answer space to 1 and 0. 
The invalid rate was 0 in the experiments, indicating that the output parsing is stable under this protocol, and the scoring is unaffected by parsing failures. 

\begin{table*}[!t]
  \fontsize{10pt}{10pt}\selectfont  
  \centering
  \renewcommand{\arraystretch}{1.2}
  \begin{tabular}{c |c| c c c c c c}
    \toprule
    \textbf{Model} & \textbf{Overall} & \textbf{LR} & \textbf{AR} & \textbf{RR} & \textbf{FP-S} & \textbf{FP-C} & \textbf{CP} \\[0.5ex]
    \midrule
    Base & 88.50 & 84.47 & 88.41 & 85.17 & 91.87 & 84.80 & 89.72\\[0.5ex]
    \textbf{FTibVLM(ours)} & 88.15 & 83.50 & 87.12 & 84.04 & 91.42 & 84.05 & 90.80\\[0.5ex]
    \bottomrule
  \end{tabular}
  \caption{
    Chinese capability retention on MMBench-CN after Tibetan adaptation.
  }
  \label{table5}
\end{table*}

\begin{table}[!t]
  \fontsize{10pt}{10pt}\selectfont  
  \centering
  \resizebox{\columnwidth}{!}{
    \begin{tabular}{c c c c c}
      \toprule
      \multirow{2}{*}{\textbf{Model}} &
      \multirow{2}{*}{\textbf{MMBench}} &
      \multicolumn{3}{c}{\textbf{POPE}} \\[0.2ex]
      & & \textbf{random} & \textbf{popular} & \textbf{adversarial} \\[0.5ex]
      \midrule
      Base & 42.97 & 47.53 & 46.38 & 46.22 \\[0.5ex]
      CP + MA & 60.87 & 71.10 & 73.83 & 69.35 \\[0.5ex]
      FTibVLM & 67.78 & 80.56 & 81.70 & 78.63 \\[0.5ex]
      \bottomrule
    \end{tabular}
  }
  \caption{\label{table4}
    Stage-wise ablation on MMBench and POPE for Tibetan adaptation (base. vs. caption alignment. vs. + instruction SFT).
  }
\end{table}

\subsection{Experimental Results} 
Tables~\ref{table1} to~\ref{table3} summarize the systematic evaluation results on FTibBench in the Tibetan setting. 
Overall, FTibVLM achieves substantial improvements over Base on POPE, BinaryVQA, COREVQA, MME, and MMBench. 

As shown in Table~\ref{table1}, on POPE, which focuses on diagnosing object hallucination, FTibVLM consistently outperforms Base across all three subsets. 
Specifically, on POPE-random, accuracy increases from 47.53 to 80.56, and F1 increases from 43.62 to 80.51. 
On the more challenging POPE-popular and POPE-adversarial subsets, accuracy reaches 81.70 and 78.63, and F1 reaches 73.40 and 78.49. 
These results indicate that after Tibetan-side adaptation, the model is more robust under the image-consistency and hallucination-sensitive conditions captured by POPE, and the gains remain consistent across subsets of different difficulty levels. 

On BinaryVQA, FTibVLM also delivers consistent gains. 
Accuracy increases from 54.46 to 76.01, and F1 increases from 53.08 to 73.25, indicating that the model’s discriminative ability is substantially strengthened in the binary VQA setting with a constrained answer space. 
For COREVQA, which emphasizes fine-grained observation and reasoning in crowded scenes, accuracy improves from 31.49 to 50.85. 

Table~\ref{table2} reports the comparison results on MME, a multi-task evaluation organized by capability dimensions. 
Overall, FTibVLM improves both accuracy and the stricter accuracy plus metric across multiple subtasks. 
For example, on basic perception tasks such as existence and color, accuracy reaches 88.33 and 78.33, while accuracy plus reaches 80.00 and 63.33. 
On tasks closer to scene understanding such as posters and scene, FTibVLM also achieves steady gains, with scene improving from 60.25 to 75.75. 
On the more challenging count task, accuracy rises from 50.00 to 66.70 and accuracy plus rises from 0.00 to 33.33. 
A finer-grained MME subtask analysis is reported in Appendix C (Table \ref{tab:mme}), which shows that improvements are most pronounced on basic perception and decision-oriented dimensions, while OCR- and text-related subtasks remain comparatively challenging. 
Given that OCR appears to be a primary bottleneck, we further conduct a targeted Tibetan OCR adaptation study; details are provided in Appendix \ref{app:ocr}.


FTibVLM increases the overall score from 42.97 to 67.78, demonstrating a substantial gain on the Tibetan multiple-choice comprehensive evaluation. 
Breaking down by dimensions, the model achieves 61.65, 63.23, and 66.07 on LR, AR, and RR, and 65.16, 68.29, and 76.01 on FP-S, FP-C, and CP. 
These results indicate that the multi-dimensional capability categories covered by MMBench reliably capture the overall improvements of the model in Tibetan multimodal understanding and reasoning. 
To complement the coverage of the main evaluation on cross-modal semantic consistency and visual entailment reasoning, we conduct an additional diagnostic study on SNLI-VE; the experimental setup and full results are reported in Appendix~\ref{app:D}.

Overall, the experimental results show that the unified evaluation protocol and hierarchical quality-control pipeline established by FTibBench can reliably differentiate Tibetan multimodal capabilities across models under the same setting. 
Within this evaluation loop, FTibVLM exhibits consistent and substantial improvements over Base across the core tasks in FTibBench, demonstrating that our stage-wise adaptation driven by FTibData effectively enhances Tibetan multimodal understanding and reasoning, and provides a reproducible and diagnosable strong baseline for future work. 

\subsection{Ablation Studies}
\subsubsection{Stage-Wise Ablation}
To quantify the marginal contributions of different data modules in FTibData and conduct an interpretable comparison under the unified evaluation protocol established by FTibBench, we evaluate three stage-wise checkpoints on FTibBench: \textbf{Base}, \textbf{CP+MA} (after continual pretraining and multimodal alignment), and the final model \textbf{FTibVLM} (further incorporating multimodal instruction tuning on top of CP+MA). 
As shown in Table~\ref{table4}, overall performance exhibits a consistent upward trend as modules are introduced, with larger gains from Base to CP+MA and additional robust improvements from CP+MA to FTibVLM. 

This stage-wise improvement aligns with the functional division of the three supervision signals. 
CP establishes a Tibetan language-distribution foundation while reducing cross-lingual forgetting, MA strengthens the alignment between visual semantics and Tibetan expressions to improve cross-modal consistency and discriminative stability, and MIT further boosts performance in interactive and multi-task settings by shaping instruction following and overall capability. 
Overall, these controlled results support a resource-module-driven and interpretable improvement conclusion: under a unified evaluation protocol and a fixed implementation setup, performance gains emerge steadily as modules are added, and can be further attributed to the incremental contributions of different modules across diagnostic dimensions of the benchmarks. 

\subsubsection{Chinese Capability Retention}
Stage-wise adaptation can inject target-language capability, but it may also introduce cross-lingual degradation. 
To examine whether our training pipeline affects Chinese multimodal performance, we compare our Tibetan multimodal model FTibVLM with the baseline Base on MMBench-CN (dev). 
As shown in Table~\ref{table5}, FTibVLM achieves an overall score of 88.15, which is essentially on par with Base at 88.50, exhibiting only minor fluctuations and no consistent downward trend. 
Notably, FTibVLM even improves on the CP dimension, increasing from 89.72 to 90.80.

These results validate that the mixed-corpus design of FTibData is both effective and necessary. 
We retain a certain proportion of Chinese in the text corpus and introduce a Chinese anchor subset in the image–text alignment and instruction fine-tuning stages, with the goal of providing cross-lingual stability constraints during training and mitigating cross-lingual forgetting and output degradation caused by continual pretraining and subsequent multi-stage adaptation. 
Overall, while substantially strengthening Tibetan multimodal capability, FTibVLM does not exhibit a noticeable loss in overall Chinese multimodal competence, providing a more stable capability boundary for cross-lingual reuse and real-world deployment in Tibetan scenarios. 
\section{Conclusion}
In this paper, we introduce FTibSuite, a resource suite for Tibetan vision--language modeling that integrates FTibData, FTibBench, and the first Tibetan VLM baseline FTibVLM, together with a reproducible training and evaluation pipeline built upon Qwen3-VL-8B-Instruct. We construct FTibData and adopt a three-stage adaptation pipeline---Tibetan continual pretraining, image--text alignment, and multimodal instruction tuning---to equip FTibVLM with Tibetan generation, grounding, and instruction-following abilities in a reproducible manner. We build FTibBench by migrating five established multimodal benchmarks into the Tibetan setting, covering hallucination robustness, binary decision stability, dense-scene understanding, capability profiling, and multiple-choice reasoning. To improve benchmark reliability, we use DeepSeek-V3 for automatic verification and rubric-based scoring, and conduct Tibetan-expert review and annotation for high-risk cases, helping partially fill the infrastructure gap for Tibetan multimodal research. Experiments show consistent Tibetan gains with minimal degradation of the backbone's Chinese capability, and staged checkpoints support controlled analysis of how different adaptation signals contribute to improvements. Future work will focus on improving Tibetan multimodal supervision quality, expanding benchmark coverage, and developing more robust multilingual adaptation strategies, especially for OCR and in-image text understanding.

\section*{Limitations}
This work focuses on data- and training-driven adaptation on top of a strong open-source backbone, rather than proposing new model architectures, so the improvements are bounded by the capabilities of the underlying design. 
In addition, the current training pipeline still has room to improve: some Tibetan multimodal supervision is obtained through translation and dataset repurposing, which may introduce noise and limit robustness. 
Future work could strengthen these aspects with higher-quality Tibetan-native multimodal data and more principled multilingual adaptation strategies. 

\section*{Ethical Considerations}

This work aims to promote inclusive vision–language modeling by extending Tibetan multimodal research through a resource suite that supports more reproducible training and evaluation. The resources in FTibSuite are constructed by adapting existing datasets and benchmark designs; we make efforts to respect original licenses and document provenance and usage constraints for each component. To improve benchmark reliability, we employ a tiered quality-control workflow with large-model automatic verification and scoring followed by Tibetan-expert review and annotation for high-risk cases. While these procedures reduce common adaptation errors and evaluation noise, residual artifacts and pretrained biases may persist, and benchmark scores should not be taken as complete evidence of real-world Tibetan multimodal competence. Finally, stronger Tibetan VLM capability may be misused, such as for generating misleading content, and we therefore encourage transparent reporting of limitations and careful deployment in high-stakes settings.

\section*{Acknowledgments}
This work was supported by the Hainan Provincial Joint Project of the Li'an International Education Innovation Pilot Zone (Grant No.~624LALH006).


\bibliography{custom}

\clearpage
\appendix
\section{FTibBench benchmark adaptation and quality control details}
\label{app:FTibBench}
\subsection*{A.1 Benchmark Composition, Splits, and Scale}

FTibBench currently includes Tibetan versions of five representative multimodal evaluation benchmarks, covering complementary dimensions such as hallucination robustness, binary decision making, dense-scene understanding, multi-dimensional capability profiling, and comprehensive multiple-choice understanding. 
Table~\ref{tab:ftibbench_mapping} summarizes the relationship between each FTibBench subset and its source benchmark, including the original language and the evaluation split used in our experiments.

\paragraph{POPE.}
POPE contains three splits (adversarial, popular, and random) and is used to evaluate hallucination tendencies and robustness under question answering about target existence. 

\paragraph{BinaryVQA.}
BinaryVQA is a binary-classification VQA benchmark whose answer space is strictly 0/1, and is used to evaluate decision stability and output controllability. 

\paragraph{COREVQA.}
COREVQA targets fine-grained observation and reasoning in dense/complex scenes, emphasizing counting, relations, and local entity understanding. 

\paragraph{MME.}
MME is a multi-dimensional capability profiling benchmark, used to diagnose a model's capability structure across multiple task dimensions. It includes 14 category subsets: artwork, celebrity, code\_reasoning, color, commonsense\_reasoning, count, existence, landmark, numerical\_calculation, OCR, position, posters, scene, and text\_translation. 

\paragraph{MMBench.}
MMBench is a multiple-choice benchmark for overall comparison of multimodal understanding and reasoning abilities.

\begin{table*}[t]
\centering
\small
\begin{tabular}{llll}
\toprule
\textbf{FTibBench subset} & \textbf{Source benchmark} & \textbf{Original language} & \textbf{Split used} \\
\midrule
FTibBench-POPE & POPE & English & Official random / popular / adversarial \\
FTibBench-BinaryVQA & BinaryVQA & English & Official split \\
FTibBench-COREVQA & COREVQA & English & Official split \\
FTibBench-MME & MME & English & Official split \\
FTibBench-MMBench & MMBench & Chinese/English & Dev \\
\bottomrule
\end{tabular}
\caption{Mapping between FTibBench subsets and their source benchmarks.}
\label{tab:ftibbench_mapping}
\end{table*}

\subsection*{A.2 Translation and Adaptation Principles}
To maximize fairness in cross-model comparisons, we follow the principle of \emph{``unchanged task definition, unchanged answer space, and structure aligned as much as possible''} during translation and adaptation. 
We strictly maintain answer-space consistency: for binary tasks, we uniformly use 0/1 (1 denotes ``yes/present/true,'' and 0 denotes ``no/absent/false''); for multiple-choice tasks, we keep the original option set unchanged and restrict outputs to A/B/C/D. 
Meanwhile, we ensure that structural fields are traceably mappable, facilitating subsequent alignment analyses and audits.

\subsection*{A.3 High-Risk Error Types and Checklist}
As is shown in Table~\ref{tab:error_taxonomy}, we treat the following error types as high risk and prioritize them for automated screening and manual review. 

\subsection*{A.4 Automated Quality Control: Evaluation Rubric and Field-Level} 
We rate each sample along three dimensions including accuracy, completeness, and expression fluency, with a total score ranging from 0 to 5, and record brief diagnostic comments to support revision and regression analysis. 
The scoring dimensions are listed in the Table~\ref{tab:translation_rubric}. 

\subsection*{A.5 Score-Triggered Revision and Manual Review Strategy} 
We adopt a tiered quality-control strategy of \emph{``\textbf{automatic scoring + human fallback}''} to balance quality and cost: 

\begin{itemize}
  \item \textbf{Total $\leq 2$: mandatory revision and mandatory human review. }
  Such samples typically exhibit missing key terms, semantic drift, or clearly unnatural phrasing, which may compromise evaluation consistency and fairness. 
  They are therefore prioritized for correction and verified by human reviewers. 
  
  \item \textbf{Total $\geq 3$: entered into a spot-check review pool. }
  Issues are usually minor, such as slight verbosity, minor over-translation, or less-than-natural wording. 
  We randomly sample \textbf{10\%} from this pool for human review: annotators re-score and label the samples, and we check whether the human judgments are consistent with the model-generated scores. 
\end{itemize}

To improve the verifiability of Tibetan translation/adaptation quality, we log, for sampled BinaryVQA instances, the English question (\texttt{question\_en}), the Tibetan question (\texttt{question}), the three dimension scores (\texttt{accuracy}/\texttt{completeness}/\texttt{tibetan\_expression}), the total score (\texttt{total}, 0-5), and a brief diagnostic comment (\texttt{comment}). During human review, we cross-check the automatic scoring results. 
Except for a small number of cases that require additional explanation, human judgments are largely consistent with the automatic scores.

\begin{figure*}[!t]
\begin{center}
\includegraphics[width=0.9\textwidth,keepaspectratio]{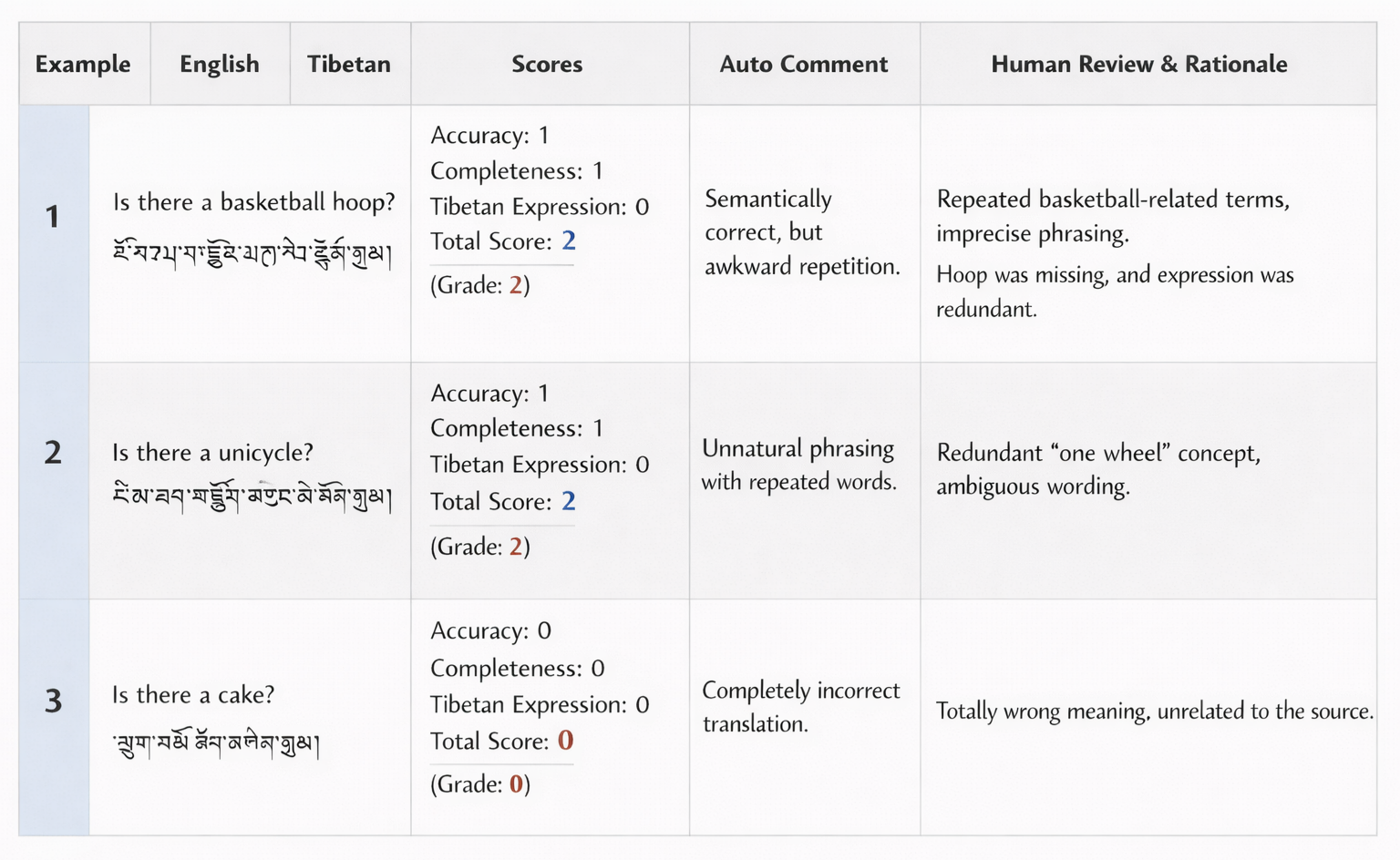}
\end{center}
\vspace{-0.4cm}
\caption{Three representative low-scoring English–Tibetan translation examples with automatic scores and human review rationales.}
\vspace{-0.3cm}
\label{fig:appendex_three_example}
\end{figure*}

As is shown in Figure~\ref{fig:appendex_three_example}, we further present three representative low-scoring examples to illustrate common translation error types and the corresponding reasons for score deductions.

\begin{table*}[t]
\centering
\small
\setlength{\tabcolsep}{4pt}
\renewcommand{\arraystretch}{1.15}
\begin{tabular}{p{0.25\linewidth} p{0.62\linewidth}}
\hline
\rule{0pt}{2.6ex}\textbf{Error Type} & \textbf{Risk Description \& Criteria} \\
\hline
\textbf{Negation Flip} &
Negation markers (e.g., not/no/non-) are dropped or weakened, resulting in a complete reversal of the intended meaning. \\
\textbf{Numerical Drift} &
Quantities, comparison relations (e.g., greater/less than), or boundary constraints (e.g., at least/at most) are altered. \\
\textbf{Entity Misalignment} &
Key entity attributes (e.g., color, position/direction, category) or proper nouns (e.g., person/place names) become inconsistent or mis-referenced. \\
\textbf{Option Mismatch} &
\emph{(Multiple-choice only)} The order of translated option texts changes, causing the correct answer label (e.g., A/B) to point to the wrong content. \\
\textbf{Option Collapse} &
\emph{(Multiple-choice only)} Distinct options are translated into near-synonymous expressions, reducing discriminability and causing overlap in the choice space. \\
\textbf{Terminology Error} &
Key objects are not translated into standardized Tibetan terminology, or neologisms are used, leading to referential ambiguity. \\
\textbf{Repetition} &
Redundant phrasing (e.g., synonym stacking or unnatural repetition of stems) appears, degrading the model's ability to parse the instruction. \\
\hline
\end{tabular}
\caption{Taxonomy of translation error types and risk description \& criteria.}
\label{tab:error_taxonomy}
\end{table*}

\begin{table*}[t]
\centering
\small
\setlength{\tabcolsep}{4pt}
\renewcommand{\arraystretch}{1.2}
\begin{tabular}{p{0.17\linewidth} p{0.07\linewidth} p{0.64\linewidth}}
\toprule
\textbf{Dimension} & \textbf{Score} & \textbf{Criteria} \\
\midrule

\multirow{3}{*}{Accuracy} 
& 2 & \textbf{Semantically accurate:} Faithfully conveys the core meaning; key information remains consistent. \\
& 1 & \textbf{Deviations present:} Largely correct, but with issues such as missing/incorrect terminology, unclear correspondences, or semantic drift. \\
& 0 & \textbf{Incorrect:} Meaning is distorted; key content is wrong or there are severe mistranslation errors. \\

\midrule

\multirow{3}{*}{Completeness}
& 2 & \textbf{Complete information:} No obvious omissions. \\
& 1 & \textbf{Minor missing:} The main message is present, but some details are missing, which may introduce slight ambiguity. \\
& 0 & \textbf{Severe missing:} Key information is missing, changing the question intent or the answer space. \\

\midrule

\multirow{2}{*}{Expression Fluency}
& 1 & \textbf{Natural:} Fluent and natural expression, with no obvious traces of literal translation. \\
& 0 & \textbf{Awkward:} Stilted or unnatural phrasing; hard to read or with clear literal-translation artifacts. \\

\bottomrule
\end{tabular}
\caption{Scoring rubric for translation quality across accuracy, completeness, and expression fluency.}
\label{tab:translation_rubric}
\end{table*}

\section{FTibVLM Three-Stage Training Setup} 
\label{app:ftibvlm} 
This section presents the end-to-end, three-stage adaptation training setup for \textbf{FTibVLM}, covering key reproducibility factors such as the training framework, hardware environment, optimizer and learning-rate schedule, LoRA configuration, batch/length settings, freezing strategy, and image resolution range.

\textbf{Stage A(CP):} LoRA $r{=}10$, $\alpha{=}16$, dropout$=0$; learning rate $5\times10^{-5}$; trained for 3 epochs; global batch size $=64$ (per-GPU batch $=1$, gradient accumulation $=8$); max sequence length $=2056$, with packing enabled; freeze the vision tower and the multimodal projector; image pixel range $[1024,\,262144]$. 

\medskip

\textbf{Stage B(CP+MA):} LoRA $r{=}10$, $\alpha{=}16$, dropout$=0$; learning rate $5\times10^{-5}$; trained for 2 epochs; global batch size $=64$ (per-GPU batch $=1$, gradient accumulation $=8$); max sequence length $=2048$ (packing disabled); freeze the vision tower while keeping the multimodal projector trainable; image pixel range $[1024,\,589824]$. 

\medskip

\textbf{Stage C(MIT):} LoRA $r{=}4$, $\alpha{=}16$, dropout$=0$; learning rate $5\times10^{-5}$; trained for 2 epochs; global batch size $=256$ (per-GPU batch $=2$, gradient accumulation $=16$); max sequence length $=2024$; gradient checkpointing enabled; freeze the vision tower while keeping the multimodal projector trainable; image pixel range $[1024,\,262144]$. 

\section{MME Subtask Evaluation Results} 
As shown in Table~\ref{tab:mme}, compared with the base model, \textbf{FTibVLM} improves \textbf{MME} \textbf{Overall Acc} from \textbf{63.98\%} to \textbf{75.02\%} (\textbf{+11.04} percentage points), and raises the stricter \textbf{Acc+} from \textbf{33.28\%} to \textbf{54.51\%} (\textbf{+21.23} percentage points). 
From a task-level breakdown, the model shows particularly pronounced gains in \texttt{existence}, \texttt{color}, and \texttt{count}, basic perception and decision-oriented tasks with clear improvements in Acc+ and output stability. 
In contrast, \texttt{OCR} and \texttt{text\_translation}, which rely more heavily on the text-recognition and cross-lingual translation pipeline, remain the main bottlenecks. 
Future work could further strengthen these capabilities by incorporating higher-quality Tibetan OCR and in-image text alignment data, as well as enforcing translation-consistency constraints.

\begin{table*}[!t]
\centering
\renewcommand{\arraystretch}{1.2}
\begin{tabular}{l|cccc}
\toprule[1.0pt]
\textbf{SubTask} & \textbf{Base(ACC)} & \textbf{Base(ACC+)}  & \textbf{FTibVLM(ACC)} & \textbf{FTibVLM(ACC+)} \\ \hline  
Existence & 50.00\% & 3.33\% & 88.33\%  &80.00\% \\
Color & 55.00\% & 10.00\% & 78.33\% &63.33\% \\ 
Code Reasoning &  62.50\% & 30.00\%  & 80.00\% & 60.00\% \\
Count & 50.00\% & 0.00\% & 66.67\% & 33.33\% \\
Artwork & 52.00\% & 8.00\% & 68.50\% & 46.00\% \\ 
Scene&  60.25\% & 31.50\% & 75.75\% & 53.00\% \\
Posters & 63.95\% & 34.69\% & 77.55\% & 59.18\% \\
Commonsense & 52.86\% & 14.29\% & 64.29\% & 34.29\% \\ 
Numerical Calc&  52.50\% & 5.00\% & 62.50\% & 25.00\% \\
Landmark & 65.50\% & 37.50\% & 72.50\% & 51.00\% \\
Position & 50.00\% & 0.00\% & 51.67\% & 20.00\% \\ 
Celebrity&  94.12\% & 88.82\% & 93.24\% & 86.47\% \\
Text Translation & 52.50\% & 10.00\% & 50.00\% & 10.00\% \\
OCR & 90.00\% & 80.00\% & 77.50\% & 55.00\% \\ 
OVERALL&  63.98\% & 33.28\% & 75.02\% & 54.51\% \\ 
 \bottomrule[1.0pt]
\end{tabular}
\vspace{-0.2cm}
\caption{The performance of FTibVLM and Qwen3-VL-8B-Instruct (Base) across MME subtasks.}
\label{tab:mme}
\end{table*}

\section{Additional Diagnostic Benchmark: SNLI-VE}
\label{app:D}
To complement the evaluation coverage of \textbf{FTibBench}, we additionally evaluate the model on the Tibetan three-way classification set of \textbf{SNLI-VE} to assess cross-modal logical consistency and visual entailment reasoning ability. 
This task requires the model to determine the relationship between an image and a textual hypothesis and output one of three labels: contradiction/neutral/entailment (encoded as 0/1/2). 
We adopt a unified scoring setup (\emph{robust candidates} + \emph{better gate}) and filter and aggregate predictions under a strict gating strategy (Gate: mode=strict\_entailment, min\_conf\_2=0.62, min\_margin\_2=0.1). 
The evaluation contains 17,901 samples, with no invalid outputs, no missing images, and no skipped samples.

As shown in Table~\ref{tab:SNLI-VE}, compared with Base, \textbf{FTibVLM} achieves a substantial improvement on this additional diagnostic task: overall \textbf{Accuracy} increases from \textbf{0.3715} to \textbf{0.5432}, and \textbf{Macro-F1} also rises from \textbf{0.3072} to \textbf{0.5400}. 
At the class level, the base model tends to over-predict \texttt{contradiction} (class \texttt{0}), whereas \textbf{FTibVLM} produces a more balanced prediction distribution and attains higher overall F1 on the \texttt{neutral} and \texttt{entailment} classes. 
These results indicate that the three-stage adaptation yields clear gains in cross-modal semantic consistency and reasoning stability.

\begin{table*}[!t]
\centering
\renewcommand{\arraystretch}{1.2}
\begin{tabular}{l|ccccc}
\toprule[1.0pt]
\textbf{Model} & \textbf{Accuracy} & \textbf{Macro-P}  & \textbf{Macro-R} & \textbf{Macro-F1} & \textbf{Prediction Distribution (0/1/2)}\\ \hline  
Base & 0.3715 & 0.3813 & 0.3716 & 0.3716 & 13626/1971/2304\\
FTibVLM & 0.5432 & 0.5703 & 0.5432 & 0.5400 & 6503/7974/3424\\ 
 \bottomrule[1.0pt]
\end{tabular}
\vspace{-0.2cm}
\caption{The performance of FTibVLM and Qwen3-VL-8B-Instruct (Base) across SNLI-VE.}
\label{tab:SNLI-VE}
\end{table*}

\section{Tibetan OCR Adaptation}
\label{app:ocr}
\subsubsection*{E.1 Motivation}
In our main experiments, we observe that although the three-stage training substantially improves Tibetan multimodal understanding and reasoning (e.g., on MME and VQA), the gains on tasks involving in-image text recognition (OCR) are not apparent. 
To examine whether \emph{language adaptation} can directly improve Tibetan OCR recognition ability, and whether a \emph{small amount of Tibetan OCR instruction data} can compensate for this capability, we conduct additional targeted experiments on Tibetan OCR. 
The results indicate that adapting a VLM to the target language, even when mixing in a small amount of OCR data into the existing instruction fine-tuning set, which does not effectively improve the model's OCR recognition ability for that language. 
OCR therefore remains one of the primary bottlenecks. 

\subsubsection*{E.2 Data and Model}
\paragraph{OCR training data.}
We collect approximately \textbf{30k} (30,000) Tibetan OCR instruction instances from our in-house resources, and mix them into the existing multimodal instruction tuning (MIT) training data for continual training, in order to test the marginal benefit of \emph{incremental OCR data}. 

\paragraph{Mixing strategy.}
We train with a \textbf{3:5} mixture ratio between the OCR data and the MIT data.

\paragraph{Compared model settings.}
We evaluate the following three model configurations on our private OCR test set: 
\begin{itemize}
  \item \textbf{FTibVLM + OCR mixed-in training:} Starting from the existing three-stage trained model, we further train by mixing \textbf{30k} OCR instruction samples into the original instruction fine-tuning data. 
  
  \item \textbf{Qwen3-VL-8B Instruct + OCR-only training on 30k:} Starting from the base instruction-tuned model, we train using \textbf{only} the \textbf{30k} OCR instruction dataset.
  
  \item \textbf{Base model (without the above OCR training):} Used as a lower-bound reference for OCR capability.
\end{itemize}

\begin{table}[!t]
  \fontsize{10pt}{10pt}\selectfont
  \centering
  \renewcommand{\arraystretch}{1.2}
  \resizebox{\columnwidth}{!}{
    \begin{tabular}{c c c}
      \toprule
      \multirow{2}{*}{\textbf{Model}} &
      \multirow{2}{*}{\textbf{CER}} &
      \multirow{2}{*}{\textbf{Exact Match}} \\[0.5ex]
      \midrule
      Base & 2.1594 & 0.0010   \\[0.5ex]
      Base + OCR-Only & 0.3283 & 0.0907 \\[0.5ex]
      CP + MA + OCR-MIT & 0.2803 & 0.3617 \\[0.5ex]
      \bottomrule
    \end{tabular}
  }
  \caption{\label{table:cer_match}
    CER and Exact Match on the Tibetan OCR benchmark for different OCR training variants.
  }
\end{table}

\subsubsection*{E.3 Experiments Results}
CER (Character Error Rate) is computed as the character-level edit distance divided by the total number of characters, where lower is better. 
Exact Match measures the proportion of samples whose predicted text matches the reference string exactly, where higher is better. 
As is shown in Figure~\ref{table:cer_match}, the results indicate a substantial gain in \emph{line-level usability} after introducing OCR supervision: the base model almost never produces correct Tibetan text on this OCR test set (Exact Match = 0.0010), while FTibVLM with mixed OCR supervision (OCR-Mix) increases Exact Match to 0.3617, i.e., an absolute improvement of +36.07 percentage points. 
This suggests that, after adding OCR data, the model can fully recognize entire text lines correctly for a considerable fraction of samples, leading to a tangible improvement in practical usability. 
In comparison, continuing training the base instruction model using only the 30k OCR dataset yields a smaller gain (Exact Match = 0.0907), and OCR-Mix achieves a clearly higher line-level accuracy. 

Despite this, CER remains relatively high, implying that OCR performance is not yet stable or uniformly reliable across samples. 
Although OCR-Mix reduces CER to 0.2803, this value still indicates non-trivial character-level errors for many instances. 
Notably, the changes in CER and Exact Match are not perfectly aligned: the large jump in Exact Match resembles a shift where a subset of samples moves from ``almost entirely wrong'' to ``entirely correct,'' rather than a uniform reduction of character errors across all samples. 
This pattern typically suggests that training primarily benefits easier sub-distributions (e.g., clear fonts, regular layouts, higher resolution, and less background clutter), while difficult cases (low resolution, occlusion, complex backgrounds, and font/style variations) still suffer from frequent character mistakes. 
Moreover, the base model sometimes exhibits CER$>1$, indicating extremely large character-level divergence from the target (e.g., many deletions/substitutions or irrelevant outputs), which is consistent with its near-zero Exact Match. 

\subsubsection*{E.4 Discussion and Implications}
This supplementary experiment suggests that \emph{language-side adaptation alone} is insufficient to obtain stable Tibetan OCR capability; improving OCR still requires \emph{dedicated supervision signals} targeting visual text recognition. 
Even with 30k OCR instruction instances, although line-level correctness increases markedly, the character error rate indicates that OCR remains far from ``stable and reliable.'' 
To systematically improve Tibetan OCR in future work, it may be necessary to incorporate: 

\begin{itemize}
  \item Larger-scale Tibetan OCR data that better matches real-world scene distributions. 
  \item Higher-resolution inputs and stronger text-region alignment supervision, e.g., line-level and box-level alignment, as well as text-region augmentation.
  \item OCR-oriented training strategies and model/component adaptations. 
\end{itemize}

\section{LLM-judge Prompt}\label{app:judge-prompt}
\subsection{LLM-judge Prompt for Translation Quality Control}

For the Tibetan translation quality-control setting, we used the prompt in Figure~\ref{fig:prompt-qc} to score an English$\rightarrow$Tibetan translation. The evaluation was conducted using DeepSeek-V3 as the LLM-judge.

\section{FTibData construction and supplementary translation-quality validation}
\label{app:FTibData}

To provide a supplementary quantitative check of translation quality, we measure semantic consistency between the original Chinese captions and their Tibetan translations on the FTibData caption subset used for multimodal alignment. Table~\ref{tab:translation_quality} summarizes the result.

\begin{table*}[t]
\centering
\small
\setlength{\tabcolsep}{10pt}
\renewcommand{\arraystretch}{1.12}
\begin{tabular}{llll}
\toprule
\textbf{Subset} & \textbf{Size} & \textbf{Metric} & \textbf{Score} \\
\midrule
FTibData-caption & 100,000 & Avg.\ cosine similarity & 0.8537 \\
\bottomrule
\end{tabular}
\caption{Supplementary quantitative validation of translation quality on the FTibData caption subset used for multimodal alignment. Semantic consistency is measured between the original Chinese captions and their Tibetan translations using a Tibetan--Chinese bilingual embedding model.}
\label{tab:translation_quality}
\end{table*}

We use this result as a lightweight validation of the semantic fidelity of the translated caption supervision, rather than as a replacement for downstream task-based evaluation or human quality control.
\begin{figure*}[ht]
\highlightbox{tred}{LLM Judge Prompt for Translation Quality Control}{
You are a strict machine-translation quality inspector. Please score the following English$\rightarrow$Tibetan translation, \textbf{strictly} following the rubric.\\

\textbf{[Scoring Dimensions]}

1) \textbf{Accuracy (0--2):} 2 = no substantive semantic errors; 1 = minor deviation; 0 = clearly misleading.\\
2) \textbf{Completeness (0--2):} 2 = no obvious omission/addition; 1 = minor omission/addition; 0 = harms the core meaning.\\
3) \textbf{Tibetan Expression (0--1):} 1 = fluent and natural; 0 = disfluent/awkward/ambiguous.\\

\textbf{[Output Requirements]}
\begin{itemize}
  \item Output \textbf{only} a single JSON object.
  \item Fields: \texttt{accuracy}, \texttt{completeness}, \texttt{tibetan\_expression}, \texttt{total}, \texttt{comment}.
  \item \texttt{total} = sum of the three scores.
  \item \texttt{comment}: a brief explanation in \textbf{English} ( $\leq$ 40 words).
\end{itemize}
}
\caption{Prompt for English$\rightarrow$Tibetan translation quality scoring.}
\label{fig:prompt-qc}
\end{figure*}

\end{document}